\documentclass[final]{cvpr}

\usepackage{times}
\usepackage{epsfig}
\usepackage{graphicx}
\usepackage{amsmath}
\usepackage{amssymb}
\usepackage{comment}
\usepackage{multirow}

\usepackage{color}

\usepackage[pagebackref=true,breaklinks=true,colorlinks,bookmarks=false]{hyperref}

\newcommand{\tablestyle}[2]{\setlength{\tabcolsep}{#1}\renewcommand{\arraystretch}{#2}\centering\small}
\newlength\savewidth\newcommand\shline{\noalign{\global\savewidth\arrayrulewidth
  \global\arrayrulewidth 1pt}\hline\noalign{\global\arrayrulewidth\savewidth}}

\def\cvprPaperID{****} 
\def\confYear{CVPR 2022}

\begin{document}

\title{Context-aware Proposal Network for Temporal Action Detection}

\author{
Xiang Wang$^{1}$
\quad Huaxin Zhang$^{1}$ 
\quad Shiwei Zhang$^{2}$
\quad Changxin Gao$^{1*}$
\quad Yuanjie Shao$^{3}$
\quad Nong Sang$^1$ 
\\
$^1${\fontsize{11.3pt}{\baselineskip}\selectfont Laboratory of Image Processing and Intelligent Control} \\
{\fontsize{11.5pt}{\baselineskip}\selectfont School of Artificial Intelligence and Automation, Huazhong University of Science and Technology} \\
$^2${\fontsize{11.5pt}{\baselineskip}\selectfont Alibaba Group}\\
$^3${\fontsize{11.3pt}{\baselineskip}\selectfont School of Electronic Information and Communications, Huazhong University of Science and Technology}\\
{\tt{\small \{wxiang, u201810644, cgao, shaoyuanjie, nsang\}@hust.edu.cn \quad zhangjin.zsw@alibaba-inc.com}}
}

\maketitle

\let\thefootnote\relax\footnotetext{${*}$ Corresponding author.}

\begin{abstract}
This technical report presents our \textcolor{blue}{\textbf{first place}} winning solution for temporal action detection task in \textbf{CVPR-2022 AcitivityNet Challenge}.
The task aims to localize temporal boundaries of  action instances with specific classes in
long untrimmed videos.
%
%
Recent mainstream attempts are based on dense boundary matchings and enumerate all possible combinations to produce proposals. 
%
We argue that the generated proposals contain rich contextual information, which may benefits detection confidence prediction. 
%
%
To this end, our method mainly consists of the following three steps:
%
%
1) action classification and feature extraction by Slowfast~\cite{slowfast}, CSN~\cite{csn}, TimeSformer~\cite{Timesformer}, TSP~\cite{tsp}, I3D-flow~\cite{i3d}, VGGish-audio~\cite{VGGish}, TPN~\cite{TPN} and ViViT~\cite{VIVIT};
2) proposal generation. Our proposed Context-aware Proposal Network (CPN) builds on top of BMN~\cite{bmn}, GTAD~\cite{gtad} and PRN~\cite{PRN} to aggregate contextual information by randomly masking some proposal features. 
%
3) action detection. The final detection prediction is calculated by assigning the proposals with corresponding video-level classification results.
Finally, we ensemble the results under different feature combination settings and achieve \textcolor{blue}{\textbf{45.8\%}} performance on the test set, which improves the champion result in CVPR-2021 ActivityNet Challenge~\cite{PRN} by \textcolor{blue}{\textbf{1.1\%}} in terms of average mAP.

\end{abstract}

\section{Introduction}

Recently, the emergence of large-scale datasets~\cite{caba2015activitynet,zhao2019hacs,liu2021multi,k700,i3d,damen2018scaling,grauman2022ego4d} and deep models~\cite{TSM,TSN,slowfast} has promoted the development of video understanding, which has a wide range of application prospects in security, surveillance, autonomous driving fields.
%
Video understanding includes many sub-research directions, such as Action Recognition~\cite{TSN,slowfast,huang2021self,hyRSM}, Action Detection~\cite{bmn,qing2021temporal,alwassel2018action,wang2021self,wang2021oadtr,wang2020cbr,wang2021weakly,MLTPN,wang2022rcl}, Spatio-Temporal Action Detection~\cite{song2019tacnet,anetava2018}, etc. 
In this report, we present our competition method for the temporal action detection task in the CVPR-2022 ActivityNet Challenge~\cite{caba2015activitynet}.

For temporal action detection task, we need to  localize temporal boundaries of action instances (\ie, start time and end time) and classify the target categories in the long untrimmed videos.
This task is challenging, involving some difficulties such as wide temporal spans of action instances, confusing background and foreground, and limited proposal contextual information.
Current mainstream approaches~\cite{bmn,gtad,tsp,wang2022rcl,xu2021low} usually adopt ``proposal and classification" paradigm, which generates proposals by calculating the boundary probabilities of each time point
to combine start points with end points and then classify the proposals.
In order to produce high-quality detection results, the generated proposals should precisely cover instance with high recalls and reliable confidence scores.
Since proposal-level classification is limited to insufficient instance information, video-level classification has attracted much attention~\cite{wang2020cbr,PRN}, which leverages the entire video as input to obtain the final results.
%
%
%
%
In this report, we follow this paradigm to design the solution of this challenge.
Our main observation is that when predicting the confidence map of dense proposals, the proposals can be mutually inferential, \ie, the confidence of a proposal may be obtained by inference from the surrounding proposals.
We thus apply a randomly masking strategy to the proposal features and encourage the model to aggregate context associations for precise proposal confidence prediction.
Moreover, to further improve the performance, we apply some data pre-processing techniques, such as too long instance removal, short instance resampling, action instance resize and temporal shift perturbation~\cite{wang2021self,TSM}.
Finally, we ensemble some existing methods~\cite{bmn,gtad,qing2021temporal,PRN} and achieve 45.8\% mAP on
the test set of ActivityNet v1.3~\cite{caba2015activitynet}, which improves the champion result in CVPR-
2021 ActivityNet Challenge~\cite{PRN} by 1.1\%.
%

%
%

%
%
\begin{figure*}[t!]
\centering
\centering{\includegraphics[width=.98\linewidth]{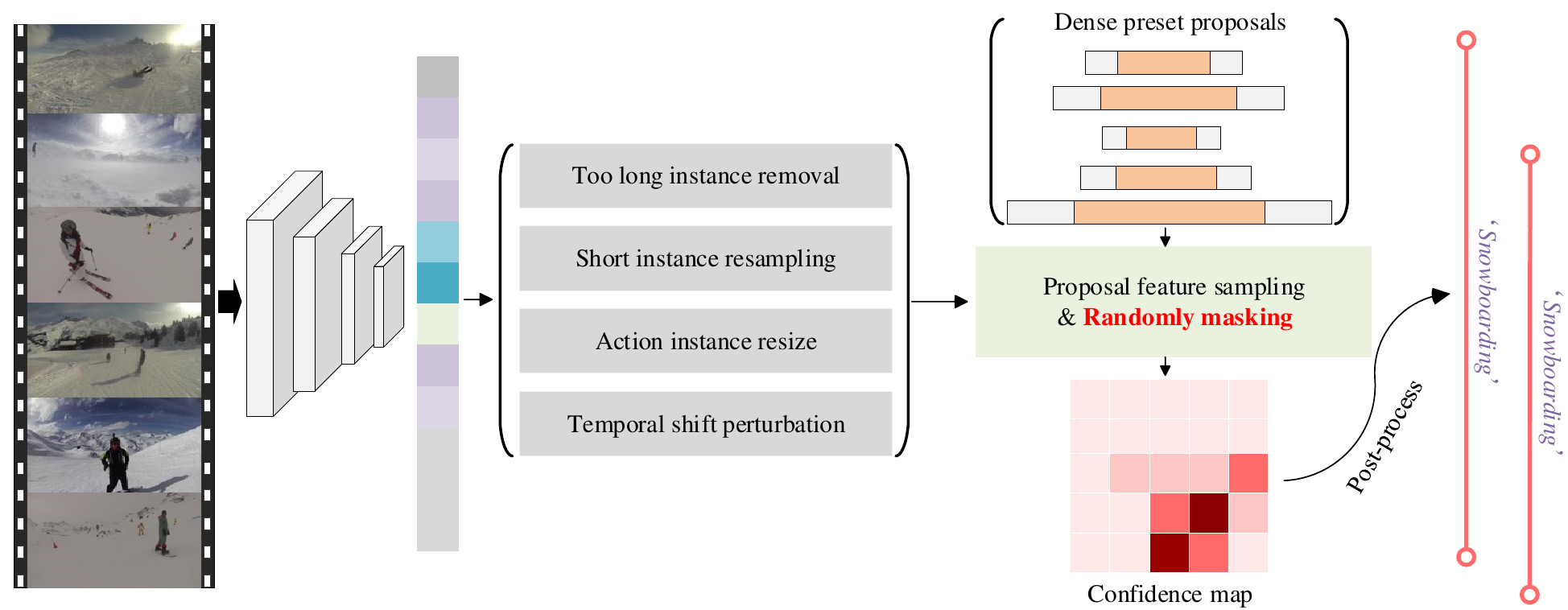}} 
\vspace{+2mm}
\caption{\label{figure-1} 
    The overall pipeline of our solution in the challenge. 
    %
    Given an input video, we utilize a pre-trained action network to obtain the feature sequence and then perform data pre-processing on the input data.
    Subsequently, we sample predefined dense proposal features on the input feature sequence and randomly drop some proposal features to encourage the use of contextual information.
    %
    Finally, post-processing the generated confidence map (including combining with video-level labels) to obtain detection results.
}
\vspace{+2mm}
\end{figure*}

\section{Feature Extractor}
In recent years, a large number of advanced deep learning algorithms have been proposed for action classification. 
%
%
These methods can act as feature extractors for action detection and also be adopted to generate video-level classifications.
%
%
In this section, we introduce some deep action classification networks used in our solution.

\subsection{Slowfast}
Slowfast network~\cite{slowfast} was proposed for action classification by combining a fast and a slow branch. 
For the slow branch, the input is with a low frame rate, which is used to capture spatial semantic information. 
The fast branch, whose input is with a high frame rate, targets to capture motion information. 
Note that the fast branch is a lightweight network, because its channel is relatively small. 
Due to its excellent performance in action recognition and detection, we choose Slowfast as one of our backbone models.

\begin{table}[t]
    \centering
\tablestyle{4pt}{1.5}
\small
\begin{tabular}{@{\;}c@{\;}|@{\;}c@{\;}|@{\;}c@{\;}|@{\;}c@{\;}}
     Model &  Pretrain  &  Top 1 Acc. &  Top5 Acc.\\
     \shline
     I3D-flow~\cite{i3d} & K400 &  79.5\% & 93.6\% \\
     Slowfast50~\cite{slowfast} & K400 &  85.3\% & 95.8\% \\
    Slowfast101~\cite{slowfast} & K400 &  87.1\% & 97.4\% \\
    Slowfast152~\cite{slowfast} & K700 &  88.9\% & 97.8\% \\
    TPN~\cite{TPN} & K400 &  87.4\% & 97.1\% \\
    TSP~\cite{tsp} & ANet & 86.4\% &  97.4\% \\
    CSN~\cite{csn} & K400 &  90.3\% & 98.1\% \\
    ViViT-B/16x2~\cite{VIVIT} & K700 &  91.2\% & 98.0\% \\
    TimeSformer~\cite{Timesformer} & K600 &  91.1\% & 97.3\% \\
    \cline{1-4}
    ANet-2020 champion~\cite{wang2020cbr} & Ensemble &  91.8\% & 98.1\% \\
    ANet-2021 champion~\cite{PRN} & Ensemble &  93.6\% & 98.5\% \\
    \textbf{Ours} & \textbf{Ensemble} &  \textbf{94.6\%} & \textbf{98.7\%} \\
    
\end{tabular}\\
\vspace{+3mm}
    \caption{Action recognition results on the validation set of ActivityNet v1.3 dataset~\cite{caba2015activitynet}. K400 means pre-training on Kinetics 400~\cite{i3d}; K700 means pre-training on Kinetics 700~\cite{k700}; ANet indicates pre-training on ActivityNet v1.3 dataset~\cite{caba2015activitynet}. Note that we also use the ActivityNet 2020 champion results~\cite{wang2020cbr} and the ActivityNet 2021 champion results~\cite{PRN} for multi-model classification fusion.}
    \label{tab:classification}
\end{table}

\subsection{I3D-flow}
Inflated 3D ConvNet (I3D)~\cite{i3d} designs some inflated convolutions to cover different receptive fields, which is based on 2D ConvNet inflation.
I3D expands the filters and pooling kernels of deep image recognition networks into 3D shape, making it suitable for spatio-temporal modeling.
In our solution, we apply I3D network to extract flow features of ActivityNet v1.3 dataset. 

\subsection{CSN}
Channel-Separated Convolutional Network (CSN)~\cite{csn} aims to reduce the parameters of 3D convolution, and extract useful information by finding important channels simultaneously.
It can efficiently learn feature representation through grouping convolution and channel interaction, and reach a good balance between effectiveness and efficiency.

\subsection{TimeSformer}
Timesformer~\cite{Timesformer} presents the standard Transformer architecture to video by enabling spatiotemporal feature learning directly from a sequence of frame-level patches. 
In addition, Timesformer shows that separate temporal attention and spatial attention within each block leads to the best video
classification accuracy.

\subsection{TPN}
Temporal Pyramid Network (TPN)~\cite{TPN} is a feature pyramid architecture, which captures the visual tempos of action instances. 
TPN can be applied to existing 2D/3D architectures in the plug-and-play manner, bringing consistent improvements. 
Considering its excellent spatio-temporal modeling ability, we also use it to extract spatio-temporal features.

\subsection{ViViT}
Due to transformers~\cite{transformer,dosovitskiy2020image,wang2021oadtr} have shown powerful abilities on various vision tasks, we apply the ViViT~\cite{VIVIT} as one of backbones.
ViViT is a pure Transformer based model for action recognition.
It extracts spatio-temporal tokens from input videos, and then encoded by series of Transformer layers.
In order to handle the long sequences of tokens encountered in videos, several efficient variants of ViViT decompose the spatial- and temporal-dimensions of the inputs.
We apply the ViViT-B/16x2 version with factorised encoder, which initialized from imagenet pretrained ViT~\cite{dosovitskiy2020image}, and then pretrain it on Kinetics700 dataset~\cite{k700}.
%


\subsection{Classification results}

In addition to the several models mentioned above, we also utilize TSP features~\cite{tsp} and VGGish-audio features~\cite{VGGish}.
Table~\ref{tab:classification} shows the action recognition results of the above methods on the validation set of ActivityNet v1.3 dataset~\cite{caba2015activitynet}. 
From the results, we can draw several following conclusions:
1) CSN model can outperform slowfast101 by 3.1\% with Kinetics400 pretraining on ActivityNet dataset;
2) Transformer based model can indeed obtain better performance than CNN based models.
For instance, TimeSformer and ViViT achieve 91.2\% and 91.1\% Top1 accuracy. 
3) The flow feature alone is not as good as the spatio-temporal feature of RGB in performance.
We then ensemble all the models and achieve 1.0\% performance gain over ActivityNet-2021 champion result.

%
%
\begin{table}[t]
    \centering
\tablestyle{4pt}{1.5}
\small
\begin{tabular}{@{\;}@{\;}c@{\;}@{\;}|@{\;}@{\;}@{\;}c@{\;}@{\;}@{\;}|@{\;}@{\;}@{\;}c@{\;}@{\;}@{\;}|@{\;}@{\;}@{\;}c@{\;}}
     Model &  Feature  &  AR@100 &  AUC \\
     \shline
    BMN~\cite{bmn} & Slowfast101 &  75.8\% & 68.6\% \\
    {PRN~\cite{PRN}} & {Slowfast101} &  {76.5\%} & {69.3\%} \\
    \textbf{CPN} & \textbf{Slowfast101} &  \textbf{76.9\%} & \textbf{69.5\%} \\
    
\end{tabular}\\
\vspace{+1mm}
    \caption{Proposal performances on the validation set of ActivityNet v1.3.}
    \label{tab:proposal}
\end{table}

\begin{table}[t]
    \centering
\tablestyle{4pt}{1.5}
\small
\begin{tabular}{@{\;}@{\;}c@{\;}@{\;}|@{\;}@{\;}@{\;}c@{\;}@{\;}@{\;}|@{\;}@{\;}@{\;}c@{\;}@{\;}@{\;}|@{\;}@{\;}@{\;}c@{\;}}
     Model &  Feature  &  0.5 &  Average mAP\\
     \shline
    BMN~\cite{bmn} & Slowfast101 &  56.3\% & 37.7\% \\
    PRN~\cite{PRN} & Slowfast101 &  57.2\% & 38.8\% \\
    CPN & Slowfast101 &  57.8\% & 39.0\% \\
    \cline{1-4}
    BMN~\cite{bmn} & Slowfast152 &  55.5\% & 36.8\% \\
    PRN~\cite{PRN} & Slowfast152 &  56.5\% & 38.0\% \\
    CPN & Slowfast152 &  56.6\% & 38.8\% \\
    \cline{1-4}
    BMN~\cite{bmn} & CSN &  56.9\% & 38.1\% \\
    PRN~\cite{PRN} & CSN &  57.9\% & 39.4\% \\
    CPN & CSN &  58.6\% & 39.5\% \\
    \cline{1-4}
    BMN~\cite{bmn} & ViViT &  55.1\% & 36.7\% \\
    PRN~\cite{PRN} & ViViT &  55.5\% & 37.5\% \\
    CPN & ViViT &  56.3\% & 38.1\% \\
    \cline{1-4}
    PRN~\cite{PRN} & Ensemble &  {59.7\%} & {42.0\%} (\textcolor{blue}{{test: 44.7\%}}) \\
     \textbf{Ours} & \textbf{Ensemble} &  \textbf{60.8\%} & \textbf{43.3\%} (\textcolor{blue}{\textbf{test: 45.8\%}}) \\
    
\end{tabular}\\
\vspace{+1mm}
    \caption{Action Detection results on the validation set of ActivityNet v1.3. Our CPN shows strong performance on multiple different features.}
    \label{tab:detection}
\end{table}

\section{Context-aware Proposal Network}

In the section, we introduce our proposed Context-aware Proposal Network (CPN).
As is shown in Figure~\ref{figure-1},
CPN mainly contains two key components: data pre-processing strategies and proposal feature random masking. 
We will introduce each part in details below, and finally show the detection performance.

\subsection{Data pre-processing strategies}
In our solution, we mainly used four data pre-processing tricks: too long instance removal, short instance resampling, action instance resize and temporal shift perturbation.

Too long instance removal means that we delete training videos where the percentage of action instances is too long (\eg, 98\%).
The intuition is that these training data lack  negative samples (\ie, the IoU between proposal and ground-truth is $0$) when generating confidence maps, which may damage the training process.

Short instance resampling denotes that the training video containing short instances is repeatedly sampled, because the recall and localization of the short video instance is low precision, and we hope to alleviate this problem by resampling.

Action instance resize is to obtain and resize action instance by ground-truth annotations, which can simulate the change in the speed of video instances.

Temporal shift operation for action recognition is first applied in TSM~\cite{TSM}, and then applied as a kind of perturbations in  SSTAP~\cite{wang2021self} for semi-supervised learning. 
Here we reuse the perturbation as the feature augmentation. 
The temporal feature shift is a channel-shift pattern, including two operations such as forward movement and backward movement in the channel latitude of the feature map. 
This module can improve the robustness of the models.
%
%

\subsection{Proposal feature random masking}

Recall that temporal action detection is to accurately locate the boundary of the target actions.
We explore the associations among proposals to capture the contextual relationships. 
To capture contextual associations among proposals, we randomly mask some proposal features. 
Specifically, a simple dropout3d operation is composed on the sampled dense proposal feature maps.
%
%
%
%
%

To evaluate proposal, we calculate AR under different Average Number of proposals (AN), termed AR@AN (\eg, AR@100), and calculate the Area under the AR \emph{vs.} AN curve (AUC) as metrics. 
Table~\ref{tab:proposal} presents the results of BMN, PRN and CPN on the validation set of ActivityNet v1.3, which prove that CPN can outperform BMN significantly. 
Especially, our method significantly improves AUC from 68.6\% to 69.5\% by
gaining 0.9\%.
In addition, compared to PRN, our CPN also has a certain performance improvement.

\subsection{Detection results}

We follow the ``proposal + classification" pipeline to generate the final detection results. 
Mean Average Precision (mAP) is adopted as the evaluation metric of temporal action detection task. 
Average mAP with IoU thresholds $[0.5 : 0.05 : 0.95]$ is applied for this challenge.

In order to demonstrate the effectiveness of CPN, we conduct experiments with different features, as is shown in Table~\ref{tab:detection}. 
The results shows that the proposed CPN can gain 1.5\% over BMN in terms of Average mAP when Slowfast101 feature is adopted. 
Then we ensemble all the results and reach 43.3\% on the validation set and 45.8\% on the test set.
The ensemble strategies mainly contain multi-scale fusion and feature combination.
We also used the boundary refinement methods~\cite{qing2021temporal,wang2020cbr} to predict boundaries more accurately.

Moreover, we can find that the Transformer based ViViT shows very strong performance on classification task but unsatisfactory on detection task when compared with the CNN models.
%
%
The reason may be that the Transformer tends to capture global information by self-attention operation, hence it loses local information which is also important for detection task.
%
%
Meanwhile, the models perform well on action task may not achieve better performance on the detection task.
Slowfast152 exceeds Slowfast101 by 0.8\% for classification, but suffers 1.2\% drop for detection in our CPN.

%

\section{Conclusion}

In this report, we present our solution for temporal action detection task in CVPR-2022 ActivityNet Challenge.
For this task, we propose a CPN to leverage rich contextual information among proposals and apply some data pre-processing strategies for high robustness.
%
Experimental results show that CPN can outperform the baseline methods significantly.
%
%
By fusing all detection results with different backbones, we obtain 45.8\% Average mAP on the test set, which gains 1.1\% over the champion method in CVPR-2021 ActivityNet Challenge.

\section{Acknowledgment}
This work is supported by the National Natural Science Foundation of China under grant 61871435, Fundamental Research Funds for the Central Universities no.2019kfyXKJC024, by the 111 Project on Computational Intelligence and Intelligent Control under
Grant B18024.

{\small
\bibliographystyle{ieee_fullname}
\bibliography{egbib}

\begin{thebibliography}{10}\itemsep=-1pt

\bibitem{tsp}
Humam Alwassel, Silvio Giancola, and Bernard Ghanem.
\newblock Tsp: Temporally-sensitive pretraining of video encoders for
  localization tasks.
\newblock In {\em Proceedings of the IEEE/CVF International Conference on
  Computer Vision (ICCV) Workshops}, 2021.

\bibitem{alwassel2018action}
Humam Alwassel, Fabian~Caba Heilbron, and Bernard Ghanem.
\newblock Action search: Spotting actions in videos and its application to
  temporal action localization.
\newblock In {\em Proceedings of the European Conference on Computer Vision
  (ECCV)}, pages 251--266, 2018.

\bibitem{VIVIT}
Anurag Arnab, Mostafa Dehghani, Georg Heigold, Chen Sun, Mario Lu{\v{c}}i{\'c},
  and Cordelia Schmid.
\newblock Vivit: A video vision transformer.
\newblock {\em arXiv preprint arXiv:2103.15691}, 2021.

\bibitem{Timesformer}
Gedas Bertasius, Heng Wang, and Lorenzo Torresani.
\newblock Is space-time attention all you need for video understanding?
\newblock In {\em International Conference on Machine Learning}, pages
  813--824. PMLR, 2021.

\bibitem{caba2015activitynet}
Fabian Caba~Heilbron, Victor Escorcia, Bernard Ghanem, and Juan Carlos~Niebles.
\newblock Activitynet: A large-scale video benchmark for human activity
  understanding.
\newblock In {\em Proceedings of the ieee conference on computer vision and
  pattern recognition}, pages 961--970, 2015.

\bibitem{k700}
Joao Carreira, Eric Noland, Chloe Hillier, and Andrew Zisserman.
\newblock A short note on the kinetics-700 human action dataset.
\newblock {\em arXiv preprint arXiv:1907.06987}, 2019.

\bibitem{i3d}
Joao Carreira and Andrew Zisserman.
\newblock Quo vadis, action recognition? a new model and the kinetics dataset.
\newblock In {\em proceedings of the IEEE Conference on Computer Vision and
  Pattern Recognition}, pages 6299--6308, 2017.

\bibitem{damen2018scaling}
Dima Damen, Hazel Doughty, Giovanni~Maria Farinella, Sanja Fidler, Antonino
  Furnari, Evangelos Kazakos, Davide Moltisanti, Jonathan Munro, Toby Perrett,
  Will Price, et~al.
\newblock Scaling egocentric vision: The epic-kitchens dataset.
\newblock In {\em Proceedings of the European Conference on Computer Vision
  (ECCV)}, pages 720--736, 2018.

\bibitem{dosovitskiy2020image}
Alexey Dosovitskiy, Lucas Beyer, Alexander Kolesnikov, Dirk Weissenborn,
  Xiaohua Zhai, Thomas Unterthiner, Mostafa Dehghani, Matthias Minderer, Georg
  Heigold, Sylvain Gelly, et~al.
\newblock An image is worth 16x16 words: Transformers for image recognition at
  scale.
\newblock {\em arXiv preprint arXiv:2010.11929}, 2020.

\bibitem{slowfast}
Christoph Feichtenhofer, Haoqi Fan, Jitendra Malik, and Kaiming He.
\newblock Slowfast networks for video recognition.
\newblock In {\em Proceedings of the IEEE/CVF International Conference on
  Computer Vision}, pages 6202--6211, 2019.

\bibitem{VGGish}
Jort~F Gemmeke, Daniel~PW Ellis, Dylan Freedman, Aren Jansen, Wade Lawrence,
  R~Channing Moore, Manoj Plakal, and Marvin Ritter.
\newblock Audio set: An ontology and human-labeled dataset for audio events.
\newblock In {\em 2017 IEEE international conference on acoustics, speech and
  signal processing (ICASSP)}, pages 776--780. IEEE, 2017.

\bibitem{grauman2022ego4d}
Kristen Grauman, Andrew Westbury, Eugene Byrne, Zachary Chavis, Antonino
  Furnari, Rohit Girdhar, Jackson Hamburger, Hao Jiang, Miao Liu, Xingyu Liu,
  et~al.
\newblock Ego4d: Around the world in 3,000 hours of egocentric video.
\newblock In {\em Proceedings of the IEEE/CVF Conference on Computer Vision and
  Pattern Recognition}, pages 18995--19012, 2022.

\bibitem{huang2021self}
Ziyuan Huang, Shiwei Zhang, Jianwen Jiang, Mingqian Tang, Rong Jin, and Marcelo
  Ang.
\newblock Self-supervised motion learning from static images.
\newblock In {\em Proceedings of the IEEE/CVF Conference on Computer Vision and
  Pattern Recognition}, 2021.

\bibitem{anetava2018}
Jianwen Jiang, Yu Cao, Lin Song, Shiwei Zhang, Yunkai Li, Ziyao Xu, Qian Wu,
  Chuang Gan, Chi Zhang, and Gang Yu.
\newblock Human centric spatio-temporal action localization.
\newblock In {\em ActivityNet Workshop on CVPR}, 2018.

\bibitem{TSM}
Ji Lin, Chuang Gan, and Song Han.
\newblock Tsm: Temporal shift module for efficient video understanding.
\newblock In {\em Proceedings of the IEEE/CVF International Conference on
  Computer Vision}, pages 7083--7093, 2019.

\bibitem{bmn}
Tianwei Lin, Xiao Liu, Xin Li, Errui Ding, and Shilei Wen.
\newblock Bmn: Boundary-matching network for temporal action proposal
  generation.
\newblock In {\em Proceedings of the IEEE/CVF International Conference on
  Computer Vision}, pages 3889--3898, 2019.

\bibitem{liu2021multi}
Xiaolong Liu, Yao Hu, Song Bai, Fei Ding, Xiang Bai, and Philip~HS Torr.
\newblock Multi-shot temporal event localization: a benchmark.
\newblock In {\em Proceedings of the IEEE/CVF Conference on Computer Vision and
  Pattern Recognition}, pages 12596--12606, 2021.

\bibitem{qing2021temporal}
Zhiwu Qing, Haisheng Su, Weihao Gan, Dongliang Wang, Wei Wu, Xiang Wang, Yu
  Qiao, Junjie Yan, Changxin Gao, and Nong Sang.
\newblock Temporal context aggregation network for temporal action proposal
  refinement.
\newblock In {\em Proceedings of the IEEE/CVF Conference on Computer Vision and
  Pattern Recognition}, 2021.

\bibitem{song2019tacnet}
Lin Song, Shiwei Zhang, Gang Yu, and Hongbin Sun.
\newblock Tacnet: Transition-aware context network for spatio-temporal action
  detection.
\newblock In {\em Proceedings of the IEEE/CVF Conference on Computer Vision and
  Pattern Recognition}, pages 11987--11995, 2019.

\bibitem{csn}
Du Tran, Heng Wang, Lorenzo Torresani, and Matt Feiszli.
\newblock Video classification with channel-separated convolutional networks.
\newblock In {\em Proceedings of the IEEE/CVF International Conference on
  Computer Vision}, pages 5552--5561, 2019.

\bibitem{transformer}
Ashish Vaswani, Noam Shazeer, Niki Parmar, Jakob Uszkoreit, Llion Jones,
  Aidan~N Gomez, Lukasz Kaiser, and Illia Polosukhin.
\newblock Attention is all you need.
\newblock {\em arXiv preprint arXiv:1706.03762}, 2017.

\bibitem{TSN}
Limin Wang, Yuanjun Xiong, Zhe Wang, Yu Qiao, Dahua Lin, Xiaoou Tang, and Luc
  Van~Gool.
\newblock Temporal segment networks for action recognition in videos.
\newblock {\em IEEE transactions on pattern analysis and machine intelligence},
  41(11):2740--2755, 2018.

\bibitem{wang2022rcl}
Qiang Wang, Yanhao Zhang, Yun Zheng, and Pan Pan.
\newblock Rcl: Recurrent continuous localization for temporal action detection.
\newblock In {\em Proceedings of the IEEE/CVF Conference on Computer Vision and
  Pattern Recognition}, pages 13566--13575, 2022.

\bibitem{MLTPN}
Xiang Wang, Changxin Gao, Shiwei Zhang, and Nong Sang.
\newblock Multi-level temporal pyramid network for action detection.
\newblock In {\em Chinese Conference on Pattern Recognition and Computer Vision
  (PRCV)}, pages 41--54. Springer, 2020.

\bibitem{wang2020cbr}
Xiang Wang, Baiteng Ma, Zhiwu Qing, Yongpeng Sang, Changxin Gao, Shiwei Zhang,
  and Nong Sang.
\newblock Cbr-net: Cascade boundary refinement network for action detection:
  Submission to activitynet challenge 2020 (task 1).
\newblock {\em arXiv preprint arXiv:2006.07526}, 2020.

\bibitem{PRN}
Xiang Wang, Zhiwu Qing, Ziyuan Huang, Yutong Feng, Shiwei Zhang, Jianwen Jiang,
  Mingqian Tang, Changxin Gao, and Nong Sang.
\newblock Proposal relation network for temporal action detection.
\newblock {\em arXiv preprint arXiv:2106.11812}, 2021.

\bibitem{wang2021weakly}
Xiang Wang, Zhiwu Qing, Ziyuan Huang, Yutong Feng, Shiwei Zhang, Jianwen Jiang,
  Mingqian Tang, Yuanjie Shao, and Nong Sang.
\newblock Weakly-supervised temporal action localization through local-global
  background modeling.
\newblock {\em arXiv preprint arXiv:2106.11811}, 2021.

\bibitem{wang2021self}
Xiang Wang, Shiwei Zhang, Zhiwu Qing, Yuanjie Shao, Changxin Gao, and Nong
  Sang.
\newblock Self-supervised learning for semi-supervised temporal action
  proposal.
\newblock In {\em Proceedings of the IEEE/CVF Conference on Computer Vision and
  Pattern Recognition}, 2021.

\bibitem{wang2021oadtr}
Xiang Wang, Shiwei Zhang, Zhiwu Qing, Yuanjie Shao, Zhengrong Zuo, Changxin
  Gao, and Nong Sang.
\newblock Oadtr: Online action detection with transformers.
\newblock In {\em Proceedings of the IEEE/CVF International Conference on
  Computer Vision}, pages 7565--7575, 2021.

\bibitem{hyRSM}
Xiang Wang, Shiwei Zhang, Zhiwu Qing, Mingqian Tang, Zhengrong Zuo, Changxin
  Gao, Rong Jin, and Nong Sang.
\newblock Hybrid relation guided set matching for few-shot action recognition.
\newblock In {\em Proceedings of the IEEE/CVF Conference on Computer Vision and
  Pattern Recognition}, pages 19948--19957, 2022.

\bibitem{xu2021low}
Mengmeng Xu, Juan~Manuel Perez~Rua, Xiatian Zhu, Bernard Ghanem, and Brais
  Martinez.
\newblock Low-fidelity video encoder optimization for temporal action
  localization.
\newblock {\em Advances in Neural Information Processing Systems}, 34, 2021.

\bibitem{gtad}
Mengmeng Xu, Chen Zhao, David~S Rojas, Ali Thabet, and Bernard Ghanem.
\newblock G-tad: Sub-graph localization for temporal action detection.
\newblock In {\em Proceedings of the IEEE/CVF Conference on Computer Vision and
  Pattern Recognition}, pages 10156--10165, 2020.

\bibitem{TPN}
Ceyuan Yang, Yinghao Xu, Jianping Shi, Bo Dai, and Bolei Zhou.
\newblock Temporal pyramid network for action recognition.
\newblock In {\em Proceedings of the IEEE/CVF Conference on Computer Vision and
  Pattern Recognition}, pages 591--600, 2020.

\bibitem{zhao2019hacs}
Hang Zhao, Antonio Torralba, Lorenzo Torresani, and Zhicheng Yan.
\newblock Hacs: Human action clips and segments dataset for recognition and
  temporal localization.
\newblock In {\em Proceedings of the IEEE/CVF International Conference on
  Computer Vision}, pages 8668--8678, 2019.

\end{thebibliography}
}

\end{document}